\documentclass[letterpaper, 10 pt, conference]{ieeeconf}  

\IEEEoverridecommandlockouts                              

\usepackage{graphics} 
\usepackage{epsfig} 
\usepackage{amsmath} 
\usepackage{amssymb}  
\usepackage{xcolor}
\usepackage{multirow}
\usepackage{booktabs}
\usepackage{hyperref}
\usepackage{cleveref}
\usepackage{pifont}
\usepackage{utfsym}

\title{\LARGE \bf
MiniVLN: Efficient Vision-and-Language Navigation by \\ Progressive Knowledge Distillation
}

\author{Junyou Zhu$^{1}$, Yanyuan Qiao$^{2}$, Siqi Zhang$^{3}$, Xingjian He$^{1}$, Qi Wu$^{2}$  and Jing Liu$^{1*}$
\thanks{*Corresponding Author: Jing Liu. E-mail:jliu@nlpr.ia.ac.cn}
\thanks{$^{1}$Junyou Zhu, Xingjian He and Jing Liu are with the Institute of Automation, Chinese Academy of Sciences, Beijing 100190, China, while
Junyou Zhu and Jing Liu are also with the School of Artificial Intelligence,
University of Chinese Academy of Science, Beijing 100190, China.}%
\thanks{$^{2}$Yanyuan Qiao and Qi Wu are with the Australian Institute for Machine Learning, The University of Adelaide, Adelaide, SA 5005, Australia.}%
\thanks{$^{3}$Siqi Zhang is with the Department of Computer Science and Technology, Tongji University, Shanghai 201804, China.}%
}

\begin{document}

\maketitle
\thispagestyle{empty}
\pagestyle{empty}

\begin{abstract}

In recent years, Embodied Artificial Intelligence (Embodied AI) has advanced rapidly, yet the increasing size of models conflicts with the limited computational capabilities of Embodied AI platforms. To address this challenge, we aim to achieve both high model performance and practical deployability. Specifically, we focus on Vision-and-Language Navigation (VLN), a core task in Embodied AI. This paper introduces a two-stage knowledge distillation framework, producing a student model, MiniVLN, and showcasing the significant potential of distillation techniques in developing lightweight models. The proposed method aims to capture fine-grained knowledge during the pretraining phase and navigation-specific knowledge during the fine-tuning phase. Our findings indicate that the two-stage distillation approach is more effective in narrowing the performance gap between the teacher model and the student model compared to single-stage distillation. On the public R2R and REVERIE benchmarks, MiniVLN achieves performance on par with the teacher model while having only about 12\% of the teacher model's parameter count.

\end{abstract}

\section{INTRODUCTION}

In recent years, there have been significant advancements in Embodied Artificial Intelligence (Embodied AI), which emphasizes agents' proficiency in interacting with their surroundings, combining multimodal information such as perception, language understanding, and action planning to complete complex tasks. Among the various tasks in Embodied AI, Vision-and-Language Navigation (VLN)~\cite{anderson2018vision} has emerged as a key research area, which requires agents to understand natural language instructions and autonomously navigate complex environments to reach target locations. 

Existing VLN methods have made substantial progress by leveraging large-scale pre-trained models to interpret multimodal information and guide agents through complex environments~\cite{chen2022duet,chen2022autovln,wang2023scalevln}. DUET~\cite{chen2022duet} integrates long-term action planning with fine-grained cross-modal understanding. AutoVLN~\cite{chen2022autovln} automatically generates a large-scale VLN dataset that significantly boosts model generalization. ScaleVLN~\cite{wang2023scalevln}, leveraging 1200+ environments and synthesizing 4.9 million instruction-trajectory pairs, exhibits significant improvements in generalization and achieves state-of-the-art results. However, many of these models are computationally intensive and require substantial memory and processing power, limiting their deployment in real-time or resource-constrained scenarios.

To address this issue, recent research has explored the potential of knowledge distillation (KD)~\cite{hinton2015distilling}. 
Knowledge distillation is a technique that replicates the performance of a larger ``teacher'' model by training a smaller, less complex ``student'' model. This approach can significantly reduce the computational requirements of VLN models while maintaining performance, making it a promising solution for deploying VLN systems in resource-limited settings.
The previous work~\cite{huang2023knowledge} focuses on the pre-training phase, training a lightweight pre-trained model using knowledge distillation. MAGIC~\cite{wang2024magic} proposes a Meta-Ability Knowledge Distillation framework and an Interactive Chain-of-Distillation learning strategy to facilitate the student model's knowledge acquisition during the fine-tuning stage.

\begin{figure}[t]
      \centering
      \includegraphics[width=1\linewidth]{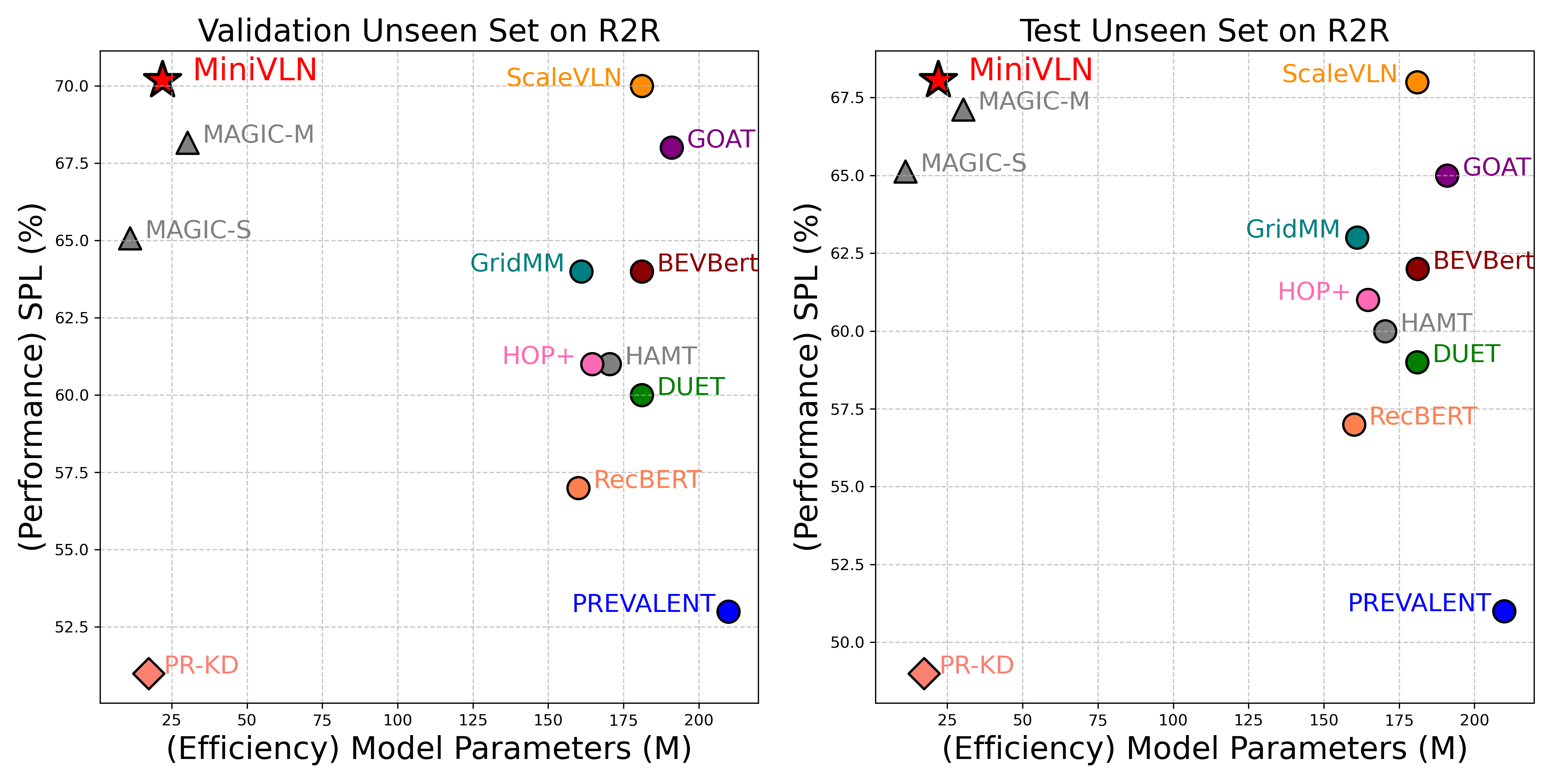}
      \caption{\textbf{Model parameters versus accuracy comparison on R2R dataset among state-of-the-art VLN methods.} Compared to other student models, MiniVLN achieves the best performance. When compared to state-of-the-art (SoTA) methods, MiniVLN uses only about 12\% of the parameters.}
      \label{fig:teaser}
   \end{figure}

In contrast to approaches~\cite{huang2023knowledge,wang2024magic} that apply distillation solely during the pre-training phase or only during the fine-tuning phase, we introduce a two-stage distillation framework, named MiniVLN. Our method incorporates knowledge distillation in both the pre-training and fine-tuning stages, leading to the final student model MiniVLN. During the pre-training stage, we focus on fine-grained knowledge learning. MiniVLN learns knowledge from the ``teacher'' model through feature alignment and representation alignment. In the fine-tuning stage, we shifted the focus to distilling knowledge that directly impacts navigation performance, such as the fused information logits used during the navigation process, which are crucial for the final navigation outcomes. To be specific, we focus on the classical dual-scale graph transformer (DUET)~\cite{chen2022duet} architecture in the VLN domain. As shown in ~\Cref{fig:teaser}, within this framework, We achieve a parameter reduction to only 12\% of the original model's size while maintaining performance comparable to that of the teacher model. Our MiniVLN surpasses the previous two student models~\cite{huang2023knowledge,wang2024magic} while keeping the model parameter count within the same order of magnitude.

This work also explores new directions for the deployment of VLN models, particularly on edge devices and other resource-limited environments. We simulate real-world deployment by running model inference on a CPU and find that MiniVLN achieves over three times the inference speed of its teacher model.

In this work, our main contributions are:
\begin{itemize}
    \item We introduce MiniVLN, a high-performance and low-complexity model specifically designed for deployment on resource-constrained devices.
    \item To the best of our knowledge, our work is the first to introduce a progressive two-stage distillation approach in the VLN field, covering both the pre-training and fine-tuning phases. Our findings indicate that two-stage distillation is more effective in bridging the performance gap between the teacher model and the student model compared to single-stage distillation.
    \item MiniVLN achieves comparable or superior performance to the teacher model on both the R2R and REVERIE datasets, despite having only 12\% of the parameters.
\end{itemize}

\section{RELATED WORK}

\subsection{Vision-and-Language Navigation}

Recently, the task of Vision-and-Language Navigation (VLN), a challenging problem in the field of Embodied-AI research, has garnered increasing attention from the research community. This area has seen substantial progress, driven by the introduction of various standard benchmarks such as Room-to-Room (R2R)~\cite{anderson2018vision}, REVERIE~\cite{qi2020reverie}, SOON~\cite{zhu2021soon}, and Room-across-Room~\cite{ku2020rxr}. 
Early VLN approaches primarily utilizes Recurrent Neural Networks (RNNs) as the backbone~\cite{anderson2018vision,fried2018speaker,vasudevan2021talk2nav,wang2019reinforced} to process sequential inputs. However, as the trajectory lengths in subsequent benchmarks increased, RNNs were found lacking in capturing long-term dependencies. 
To address the challenges of long-term memory in navigation tasks, Transformer-based models were introduced~\cite{chen2021hamt,guhur2021airbert,hao2020prevelant,chen2022duet,hong2020rvlnbert}. For instance, RecBERT~\cite{hong2020rvlnbert} utilizes the [CLS] token within the transformer as a recurrent state to record navigation history, while HAMT~\cite{chen2021hamt} efficiently models the long-horizon history of panoramic observations and actions. DUET~\cite{chen2022duet} combined long-term action planning with fine-grained cross-modal understanding, dynamically constructing a topological graph to integrate both local observation and global map. 
On this basis, advancements are achieved through data augmentation~\cite{chen2022autovln,wang2022counterfactual,wang2022less,zeng2023kefa}, external knowledge integration~\cite{li2023kerm}, and visual representation refinement~\cite{hong2023ego2map,an2023bevbert,liu2023bsg,wang2023gridmm,liu2024volumetric}.
Recently, ScaleVLN~\cite{wang2023scalevln} employed environments from HM3D~\cite{ramakrishnan2021hm3d} and Gibson~\cite{xia2018gibson} as sources of visual data, synthesizing a large number of high-quality instruction-trajectory pairs. This approach has achieved state-of-the-art performance on datasets like R2R and approached human-level performance. 
Despite the significant progress, limited attention has been devoted to addressing the challenges of model complexity and practical deployment. VLN-PETL~\cite{qiao2023vln-petl} endeavored to minimize task-specific parameters but led to an increase in computation complexity during inference due to the additional modules.
In this paper, we propose to derive a compact model through knowledge distillation, aiming to develop an efficient VLN model while maintaining competitive or even superior performance.

\subsection{Knowledge Distillation}
The goal of knowledge distillation is to transfer the knowledge learned by a large, complex teacher network to a smaller, lightweight student model. By learning from the soft labels provided by the teacher model, the student model can achieve performance levels comparable to the teacher model,  often surpassing what is possible with hard label training alone.  Knowledge distillation has been successfully applied across various computer vision tasks, including object detection\cite{cao2022pkd,wang2024crosskd}, image segmentation\cite{liu2019structuredsegkd,yang2022cross}, and pose estimation\cite{li2021online}.  TinyBERT~\cite{jiao2019tinybert} introduced distillation of the multi-head attention layer and fully connected feed-forward layer, resulting in a lightweight model distilled from the teacher model. 

Recently, attention has shifted towards distillation in the VLN domain such as \cite{huang2023knowledge}, where a transformer-based model was distilled during the pretraining stage to obtain a student model. MAGIC~\cite{wang2024magic} proposed a Meta-Ability Guided Interactive Chain-of distillation method to train the student model in the fine-tuning phase. We combined pre-training and fine-tuning phases to train the student model, which outperforms the previous two methods under the condition of having comparable model parameters.

\section{Preliminaries}
\subsection{{Problem Formulation}}
Vision-and-Language Navigation (VLN) tasks involve an agent navigating through an unseen environment based on natural language instructions. The environment is modeled as an undirected graph $\mathcal{G} = \{\mathcal{V}, \mathcal{E}\}$, where $\mathcal{V}$ represents a set of navigable nodes, and $\mathcal{V}$ denotes the connectivity edges between these nodes. The agent is initialized at a random starting node.
The goal of the agent is to interpret a given natural language instruction \( \mathcal{I} = \{w_i\}_{i=1}^L \), where \( L \) is the length of the instruction, and navigate through the connectivity graph to reach the specified target location. 
This process is formulated as a partially observable Markov decision process (POMDP), where the agent's future observations are conditionally independent of past observations given the current state \( s_t \).

At time step \( t \), the agent receives a panoramic observation \( \mathcal{O}_t = \{ o_{t,i}, a_{t,i} \}_{i=1}^K \) from its current viewpoint \( V_t \). This observation consists of a set of images capturing the surrounding environment, which are split into \( K \) individual views $o_{t,i}$ along with its associated angle direction $a_{t,i}$. 
At fine scale, the action space $\mathcal{A}_t^f$ comprises navigating to one of the neighboring nodes $\mathcal{N}(V_t)$ and stopping. At coarse scale, the action space $\mathcal{A}_t^c = \bigcup_{s=1}^t \mathcal{N}(V_s) \backslash V_s$ consists of navigating to all the navigable but unvisited nodes and stopping.
The agent must learn a policy \( \pi \) that predicts the next action based on the instruction \( \mathcal{I} \), the agent's navigation history, and the current observation \( \mathcal{O}_t \).
The agent's decision-making process continues until it chooses to stop at a location, at which point it must locate the target object within the panoramic observation. The overall goal is to optimize the agent's ability to follow the instructions accurately, navigate efficiently through the environment, and locate the target.

\subsection{Baseline Model}
\label{sec:duet}
We adopt DUET~\cite{chen2022duet} as our baseline method, which constructs a topological map $\mathbf{G}_t = \{\mathbf{N}_t, \mathbf{E}_t\}$ over time to memorize visited locations and combines coarse-scale map encoding with fine-scale encoding of the current location to enhance action planning.
$\mathbf{N}_t$ comprises visited nodes, the current node, and ghost nodes representing navigable but unvisited nodes. $\mathbf{E}_t$ records the Euclidean distances between adjacent nodes.
DUET consists of a language encoder and a panorama encoder for single-modal embedding and feature extraction, along with two cross-modal encoders for coarse- and fine-scale multi-modal fusion.

\noindent\textbf{Language Encoder}
The instruction $\mathcal{I}$ is first embedded to $\mathbf{E} = \{\hat{w}_i\}_{i=1}^L$. 
Subsequently, $\mathbf{E}$ undergoes through $N_L$ transformer blocks to generate the instruction feature $\mathbf{f}_L \in \mathbb{R}^{L\times D}$, where each layer comprises a multi-head self-attention (MSA) layer and a feed-forward network (FFN).

\noindent\textbf{Panorama Encoder}
At time step $t$, a pre-trained vision transformer (ViT)~\cite{dosovitskiy2020vit} is first applied to extract feature vectors $\mathbf{E}_v$ from panoramic observations $\{o_{t,i}\}_{i=1}^K$. $\mathbf{E}_v$ is 
input into $N_P$ transformer blocks to generate panoramic features $\mathbf{h}_t \in \mathbb{R}^{K\times D}$, which is incorporated with a type embedding and a location embedding indicating the relative position to the start node.
During the pre-training phase, the observations of navigation history $V_1,...,V_{t-1}$ are encoded in the same way.

\noindent\textbf{Coarse-scale Cross-modal Encoder}
To construct the topological map feature $\mathbf{f}_t^c \in \mathbb{R}^{|\mathbf{N}_t|\times D}$, the average of panoramic features are taken to represent visited nodes and the current node. Ghost nodes are partially observed and are represented by accumulated features from other nodes. Subsequently, $\mathbf{f}_L$ and $\mathbf{f}_t^c$ are passed through $N_X$ cross-modal transformer blocks which consist of a multi-head cross-attention layer, a MSA layer, and a FFN to generate multi-modal map features for coarse-scale action prediction.

\noindent\textbf{Fine-scale Cross-modal Encoder}
$\mathbf{f}_L$ and $h_t$ are passed through $N_X$ cross-modal transformer blocks to generate multi-modal local features for fine-scale action prediction.

\section{Method}
\begin{figure}[t]
      \centering
      \includegraphics[width=\linewidth]{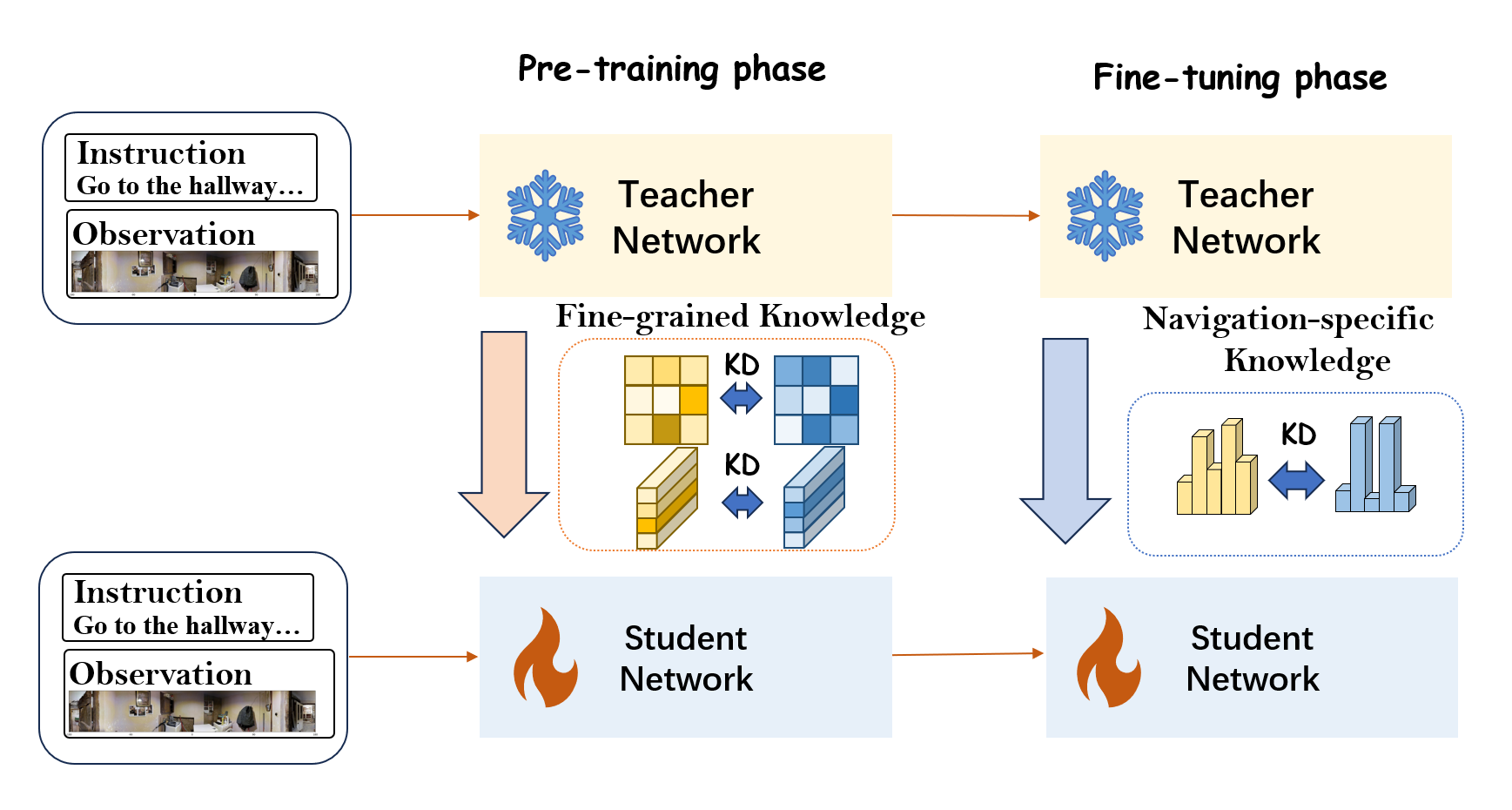}
      \caption{\textbf{The overview of two-stage knowledge distillation process for VLN. }In the pre-training phase, fine-grained knowledge is distilled, while navigation-specific knowledge is learned during fine-tuning.  This approach better narrows the performance gap between the teacher and student models compared to single-stage distillation.}
      \label{fig:overview}
   \end{figure}

Current VLN methods follow a pretraining-and-finetuning paradigm~\cite{chen2022duet,chen2022autovln,wang2023scalevln}. In the pre-training phase, the agent learns to align the features of the instruction and oracle navigation history and predict the next action based on them. While in the fine-tuning phase, the agent iteratively predicts actions according to the instruction and its actual navigation history.
On this premise, we propose MiniVLN with two distinct distillation strategies tailored for each training phase.
As illustrated in ~\Cref{fig:overview}, knowledge pertaining to feature learning and alignment is distilled in the pre-training phase, while the fine-tuning phase focuses on distilling knowledge regarding navigation action determination.

The ScaleVLN~\cite{wang2023scalevln} model, which shares the same architecture as DUET~\cite{chen2022duet} (as depicted in \Cref{sec:duet}), is employed as our teacher model. For clarity, we denote the pretraining outcome of ScaleVLN as \texttt{Scale$_{\texttt{pre}}$}, and the fine-tuning outcome as \texttt{Scale$_{\texttt{ft}}$}. They serve as the respective teacher models of our two-stage distillation process.
   
\begin{figure*}[t]
      \centering
      \includegraphics[width=1\textwidth]{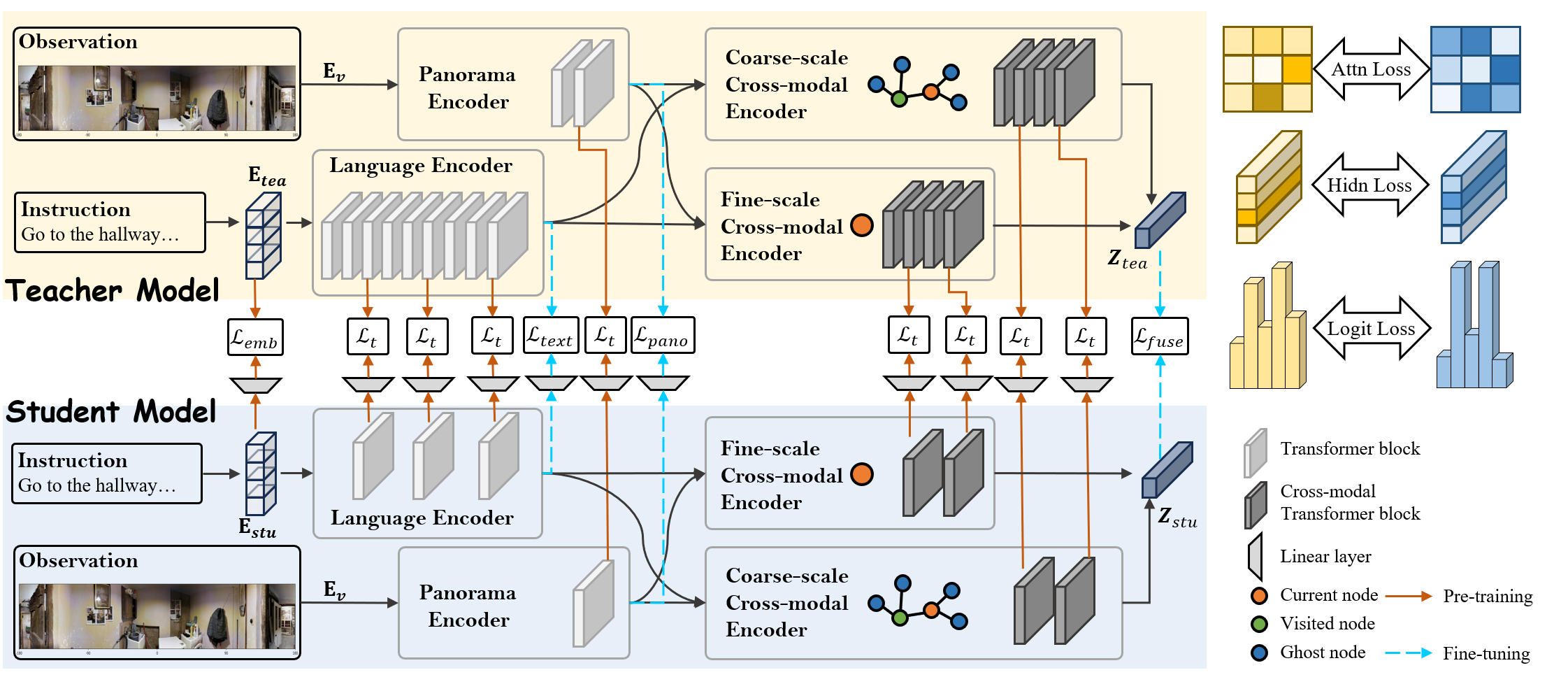}
      \caption{\textbf{
      Overall framework of MiniVLN.
      } The yellow box represents the teacher model, while the blue box denotes the student model. The orange arrows represent the distillation process during the pre-training phase, while the blue arrows denote the distillation during the fine-tuning phase. During the pre-training phase, we perform fine-grained distillation by designing $\mathcal{L}_{t}$, which distills knowledge between Transformer layers for feature and representation learning. In the fine-tuning phase, we distill only the logits directly related to navigation. }
      \label{fig:pre}
   \end{figure*}

\subsection{Knowledge Distillation During Pretraining Phase}

In order to distill knowledge encapsulated within the teacher model's learned features, we conduct Embedding Distillation, Attention-based Distillation, and Hidden States-based Distillation in the pre-training phase.

\noindent\textbf{Embedding Distillation}
involves calculating the Mean Squared Error (MSE) loss between the embedding layers of the teacher model and the student model:
\begin{equation}
    \mathcal{L}_{\text{emb}} = \texttt{MSE}(\mathbf{E}_{\text{tea}}, \mathbf{E}_{\text{stu}}\mathbf{W}_{\text{e}})
\end{equation}
where $\mathbf{E}_{\text{tea}(\text{stu})}$ are the instruction embeddings of teacher (student) models, and $\mathbf{W}_{\text{e}}$ is a learnable scaling matrix.

\noindent\textbf{Attention-based Distillation}
The core component of a transformer block is the attention layer:
\begin{equation}
    \mathbf{A} = \frac{\mathbf{X}\mathbf{W}_q(\mathbf{X}\mathbf{W}_k)^T}{\sqrt{D}}
\end{equation}
Given the crucial role of capturing interdependencies between tokens for contextual comprehension, the student model is trained to align with the attention matrix of the teacher model:
\begin{equation}
    \mathcal{L}_{attn} = \frac{1}{h}\sum_{i=1}^h \texttt{MSE}(\mathbf{A}_i^T, \mathbf{A}_i^S)
\end{equation}
where $h$ denotes the number of attention heads, and $\mathbf{A}_i^{T(S)}$ refers to the attention matrix in the $i$-th head of teacher (student) model.

\noindent\textbf{Hidden States-based Distillation} aims to enable the student model to distill knowledge from the output features of teacher model's transformer block:
\begin{equation}
    \mathcal{L}_{hidn} = \texttt{MSE}(\mathbf{H}^T, \mathbf{H}^S\mathbf{W}_h)
\end{equation}
where $\mathbf{H}^{T(S)}$ denotes the hidden states of the teacher (student) model, and $\mathbf{W}_h$ is a learnable matrix.

\noindent\textbf{Distillation Loss}
The language encoder and panorama encoder in \texttt{Scale$_{\texttt{pre}}$} consists of $N_L=9$ and $N_P=2$ transformer blocks respectively. Both the coarse- and fine-scale cross-model encoders include $N_X=4$ cross-model transformer blocks.
As the student model, our MiniVLN is much smaller, where $N_L'=3$, $N_P'=1$, and $N_X'=2$.
The mapping from student blocks to teacher blocks is illustrated in \Cref{fig:pre}, where the $m$-th block of MiniVLN learns knowledge from the $h(m)$-th block of \texttt{Scale$_{\texttt{pre}}$}. For both transformer blocks and cross-modal transformer blocks, distillation is applied to the self-attention matrix and the output features:
\begin{equation}
\small
    \mathcal{L}_{t,m} = \frac{1}{h}\sum_{i=1}^h \texttt{MSE}(\mathbf{A}_{h(m),i}^T, \mathbf{A}_{m,i}^S) + 
    \texttt{MSE}(\mathbf{H}_{h(m)}^T, \mathbf{H}_{m}^S\mathbf{W}_h)
\end{equation}
where $h(m)=3m$ in the language encoder, while $h(m)=2m$ in the panorama encoder and cross-modal encoders.

\subsection{Knowledge Distillation During Finetuning Stage}

As shown by the blue dashed lines in \Cref{fig:pre}, during the finetuning stage, three key points for knowledge distillation are designed. First, at the start of navigation, the student model needs to learn the knowledge from the teacher model when encoding the given instruction. Second, during the navigation process, at each viewpoint, the student model needs to learn the teacher model's encoding of the current panoramic visual observation. Finally, the fused information of local information and global information is distilled.

\noindent\textbf{Text Encoder Distillation} The agent begins navigation by receiving a textual instruction $I$. The instruction is encoded by the teacher and student model respectively to produce $\mathbf{f}_L^T$ and $\mathbf{f}_L^S$.
To distill the knowledge from the teacher model to the student model, the MSE between these outputs is calculated as:

\begin{equation}
\mathcal{L}_{\text{txt}} = \texttt{MSE}(\mathbf{f}_L^T, \mathbf{f}_L^S\mathbf{W}_l)
\end{equation}
where $\mathbf{W}_l$ is a learnable weight matrix that aligns the output space of the student model with that of the teacher model.

\noindent\textbf{Panorama Encoder Distillation} As the agent navigates, it receives $K=36$ images at each viewpoint. These images are averaged to form a fused panoramic observation $R$. 
The MSE loss between the outputs of the teacher and student models for this panoramic observation is computed as:
\begin{equation}
\mathcal{L}_{\text{pano}} = \texttt{MSE}(\mathbf{h}_t^T, \mathbf{h}_t^S\mathbf{W}_r)
\end{equation}
where $\mathbf{W}_r$ is another learned weight matrix that adapts the student model's representation to match that of the teacher model for the panoramic observations.

\noindent\textbf{Fusion of Local and Global Information} Finally, the global and local navigation information is combined, and the weighted sum of global and local information is calculated through a fusion strategy, ultimately producing logits that represent the probabilities of each navigation point. Let the teacher model's probabilities be $\mathbf{Z}_{tea}$ and the student model's probabilities be $\mathbf{Z}_{stu}$. The loss function is then designed as:
\begin{equation}
    \mathcal{L}_{\text{fuse}} = \texttt{CE}\left(\frac{\mathbf{Z}_{tea}}{t}, \frac{\mathbf{Z}_{stu}}{t}\right)
\end{equation}
where $\texttt{CE}$ denotes the cross-entropy loss, and $t$ represents the temperature value. In our experiment, we set $t = 1$.

Compared to TinyBERT, the distillation method proposed in this paper includes certain optimizations. We have specifically tailored local distillation, global distillation, and fusion distillation for the DUET model, and performed distillation normalization to enhance performance. Finally, the overall knowledge distillation loss $\mathcal{L}_{\text{ft\_kd}}$ is the sum of these three types of distillation losses:

\begin{equation}
\mathcal{L}_{\text{ft\_kd}} =\mathcal{L}_{\text{txt}} + \mathcal{L}_{\text{pano}} + \mathcal{L}_{\text{fuse}}
\end{equation}

\begin{table}[t]
\caption{Comparison with SoTA methods on the R2R dataset.}
\label{table1}
\begin{center}
\resizebox{\linewidth}{!}{
\begin{tabular}{l|cc|cc|cc}
\toprule
\multirow{2}{*}{\textbf{Method}}  & \multicolumn{2}{c|}{\textbf{Val Unseen}} & \multicolumn{2}{c|}{\textbf{Test Unseen}}& \multirow{2}{*}{\textbf{Param(M)}$\downarrow$}  \\ 
&\textbf{SR}$\uparrow$ & \textbf{SPL}$\uparrow$ & \textbf{SR}$\uparrow$ & \textbf{SPL}$\uparrow$ & \\
\hline
PREVALENT~\cite{hao2020prevelant}  &  57 & 53 & 54 & 51 & 209.83 \\
RecBERT~\cite{hong2020rvlnbert}  &  63 & 57 & 63 & 57 &  159.99 \\
HAMT~\cite{chen2021hamt} &  66 & 61 &  65 & 60 & 170.39 \\
ADAPT~\cite{lin2022adapt}  & 66 & 59  & 63 & 57 & 161.96  \\
DUET~\cite{chen2022duet}   & 72 & 60  & 69 & 59 & 181.02 \\
HOP+~\cite{qiao2023hop+}  & 67 & 61  & 66 & 60 & 164.62\\
TD-STP~\cite{zhao2022target}   & 70 & 63 & 67 & 61  & 171.87 \\
KERM~\cite{li2023kerm}   & 72 & 61  & 70 & 59 & 222.04  \\
GeoVLN~\cite{huo2023geovln}  & 68 & 63  & 65 & 61 & 182.64  \\
DSRG~\cite{wang2023dualDSRG}  &  73 & 62 & 72 & 61 & 188.87  \\
BEVBert~\cite{an2023bevbert}   & 75 & 64 & 73 & 62 & 181.08 \\
GridMM~\cite{wang2023gridmm}   & 75 & 64 & 73 & 62 & 161.00 \\
GOAT~\cite{wang2024visionGOAT}  & 78 & 68  & 75 & 65  & 190.96\\
ScaleVLN~\cite{wang2023scalevln} & 79 & 70 & 77 & 68  & 181.02  \\
\hline
\multicolumn{6}{c}{\textit{\textbf{Knowledge Distillation Methods}}} \\
\hline
PR-KD~\cite{huang2023knowledge} & 44.7 & 51.1 & 52 & 49.4 & 17.30 \\
MAGIC-M~\cite{wang2024magic}  & 77.39 & 68.16 & 75.53 & 67.12 & 30.23 \\
MAGIC-S~\cite{wang2024magic}  & 76.03 & 65.07 & 75.17 & 65.13 & \textbf{11.14}  \\
\hline
MiniVLN (ours) & \textbf{78.80} & \textbf{70.17} & \textbf{77.59} & \textbf{68.05} & 21.95\\
\bottomrule
\end{tabular}}
\end{center}
\end{table}

\begin{table*}[htb]
\caption{Comparison with SoTA methods on the REVERIE dataset.}
\label{table2}
\setlength{\tabcolsep}{4mm}
\centering
\begin{tabular}{l|cccc|cccc|cc}
\toprule
\multirow{2}{*}{\textbf{Method}}  & \multicolumn{4}{c|}{\textbf{Validation Unseen}} & \multicolumn{4}{c|}{\textbf{Test Unseen}}& \multirow{2}{*}{\textbf{Param(M)$\downarrow$}}  \\
   &\textbf{SR$\uparrow$} & \textbf{SPL$\uparrow$} &\textbf{RGS$\uparrow$}&\textbf{RGSPL$\uparrow$} &\textbf{SR$\uparrow$} & \textbf{SPL$\uparrow$}&\textbf{RGS$\uparrow$}&\textbf{RGSPL$\uparrow$} &  \\
\midrule

HAMT~\cite{chen2021hamt} & 32.95 & 30.20 & 18.92 & 17.28 & 30.40 & 26.67 & 14.88 & 13.08 & 170.39\\
DUET~\cite{chen2022duet} & 46.98 & 33.73 & 32.15 & 23.03 & 52.51 & 36.06 & 31.88 & 22.06 & 181.02\\ 
GridMM~\cite{wang2023gridmm} & 51.37 & 36.47 & 34.57 & 24.56 & 53.13 & 36.60 & 34.87 & 23.45 & 161.00 \\
AutoVLN~\cite{chen2022autovln} & \textbf{55.89} & 40.85
 & \textbf{36.58} & 26.76 & \textbf{55.17} & 38.88 & \textbf{32.23} & 22.68 & 181.02 \\
\hline
MiniVLN (ours)& 54.30 & \textbf{42.02} & 35.16 & \textbf{27.06} & 53.62 & \textbf{39.66} & 31.18 & \textbf{23.33} & \textbf{21.95} \\
\bottomrule
\end{tabular}
\end{table*}

\section{Experiments}
\subsection{Experimental Setup}
\subsubsection{Datasts}
ScaleVLN~\cite{wang2023scalevln} primarily focuses on the R2R dataset. Therefore, we employ ScaleVLN as the teacher model for knowledge distillation on this dataset. On the REVERIE dataset, we apply AutoVLN~\cite{chen2022autovln} as our teacher model.
{R2R} comprises 22k human-annotated navigational instructions. Each instruction, averaging 32 words in length, provides step-by-step guidance for navigation. The associated expert paths consist of six or seven nodes, covering a total distance of approximately 10 meters. This dataset offers a challenging testbed for vision-and-language navigation tasks.
{REVERIE} provides high-level instructions, averaging 21 words, focusing on target locations and objects. Agents follow 4 to 7 step trajectories and must select the correct object from predefined bounding boxes at the end of the path. This task extends R2R by adding object localization.

\subsubsection{Evaluation Metrics}
We assess agent performance using standard VLN metrics, including Success Rate (SR) and Success weighted by Path Length (SPL). Remote Grounding Success (RGS) and its path-length penalized version (RGSPL) evaluate object localization accuracy.

\begin{figure}[tbp]
      \centering
      \includegraphics[width=1\linewidth]{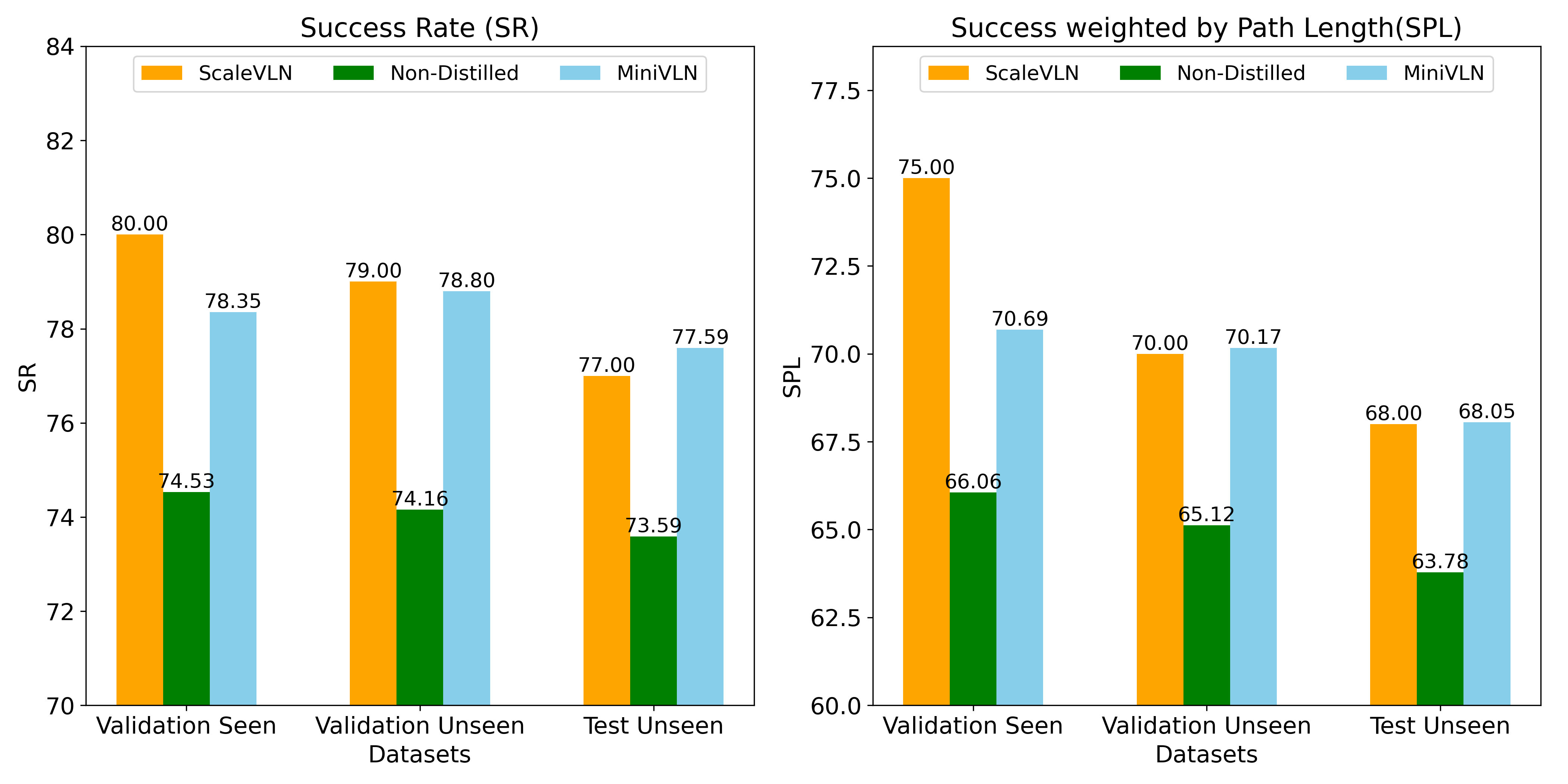}
      \caption{\textbf{Ablation of two-stage distillation on the R2R dataset.} MiniVLN maintains performance comparable to the teacher model while achieving approximately 4\% higher performance than the non-distilled model.}
      \label{fig:a1}
\end{figure}

\subsubsection{Training Details}

We trained on the R2R dataset for 200,000 iterations with a batch size of 16, and on the REVERIE dataset for 20,000 iterations with a batch size of 32. These settings were applied consistently across all experiments in this paper.

\subsection{Comparisons with State-of-the-Arts}

\subsubsection{Results on R2R}

As shown in \Cref{table1}, in the test unseen setting, MiniVLN achieves a SR of 77.59 and a SPL of 68.05, surpassing ScaleVLN~\cite{wang2023scalevln} which achieves 77 SR and 68 SPL. Additionally, the model size of MiniVLN is approximately one-ninth of that of ScaleVLN.
Compared to MAGIC-M and MAGIC-S~\cite{wang2024magic}, MiniVLN demonstrates superior performance while maintaining a comparable parameter size on both the validation unseen and test unseen sets. This highlights the efficacy of our two-stage distillation framework compared to single-stage approaches. As shown in \Cref{fig:teaser}, MiniVLN outperforms all previous state-of-the-art (SoTA) methods on the R2R dataset, with only 12\% (22M) of the parameters compared to previous models.

\subsubsection{Results on REVERIE}
\Cref{table2} compares the performance of our model, MiniVLN, with several classic methods on the REVERIE dataset. MiniVLN outperforms the teacher model in path-weighted metrics, achieving 102\% of the teacher model’s SPL and 103\% of its RGSPL. Additionally, MiniVLN attains approximately 97\% of the teacher model’s performance in SR and RGS. Notably, these results are achieved with MiniVLN being only about one-ninth the size of the models listed in \Cref{table2}.

\subsection{Ablation Study}
\subsubsection{The Effect of Two-Stage Distillation}
To demonstrate the effectiveness of our two-stage distillation process, we conduct experiments using TinyBERT with the model parameters kept constant. On the R2R datasets, the results, as shown in ~\Cref{fig:a1}, reveal that the non-distilled model achieves an SR of only 74.16 and an SPL of 65.15 on the validation unseen set, and an SR of 73.59 and an SPL of 63.78 on the test set. In contrast, our MiniVLN model achieves an SR of 78.80 and an SPL of 70.17 on the validation unseen set, and an SR of 77.59 and an SPL of 68.05 on the test set, highlighting the significant performance improvement attained through our distillation approach. Additionally, ablation experiments on the REVERIE dataset, detailed in \Cref{table4}, illustrate the contributions of each stage of the distillation process, highlighting the effectiveness of both the fine-grained knowledge distillation during pre-training phase (model$\#3$) and navigation-specific knowledge distillation during fine-tuning phase (model$\#2$). In \Cref{table4}, model$\#4$ shows further illustrates the effectiveness of our progressive two-stage approach.

\begin{table}[t]
\caption{Ablation Study on the Effect of Two-Stage Distillation on the REVERIE Dataset.}
\label{table4}
\centering
\begin{tabular}{l|cc|cccc}
\toprule
\multirow{2}{*}{Id}  & \multirow{2}{*}{Pre-train} & \multirow{2}{*}{Fine-tune} & \multicolumn{4}{c}{Validation Unseen} \\
   & & &SR & SPL &RGS&RGSPL  \\
\midrule
\#1 & \ding{51} & \ding{51} & \textbf{54.30} & \textbf{42.02} & \textbf{35.16} & \textbf{27.06} \\
\hline
\#2 & \ding{51} & \ding{55} & 54.13 & 38.64 & 31.89 & 22.93 \\
\#3 & \ding{55} & \ding{51} & 52.60 & 41.77 & 33.60 & 26.47 \\
\#4 & \ding{55} & \ding{55} & 49.16 & 36.96 & 30.99 & 23.16 \\
\bottomrule
\end{tabular}
\end{table}

\subsubsection{The Weight of Knowledge Distillation}
In the fine-tuning phase, the overall loss comprises both the original action classification loss and the knowledge distillation loss $\mathcal{L}_{\text{ft\_kd}}$. The weight of $\mathcal{L}_{\text{ft\_kd}}$, referred to as KDWeight, is ablated in \Cref{table6}.
The results show that a KDWeight of 0.1 performs best on the validation seen set and validation unseen set.

\begin{figure}[t]
      \centering
      \includegraphics[width=1\linewidth]{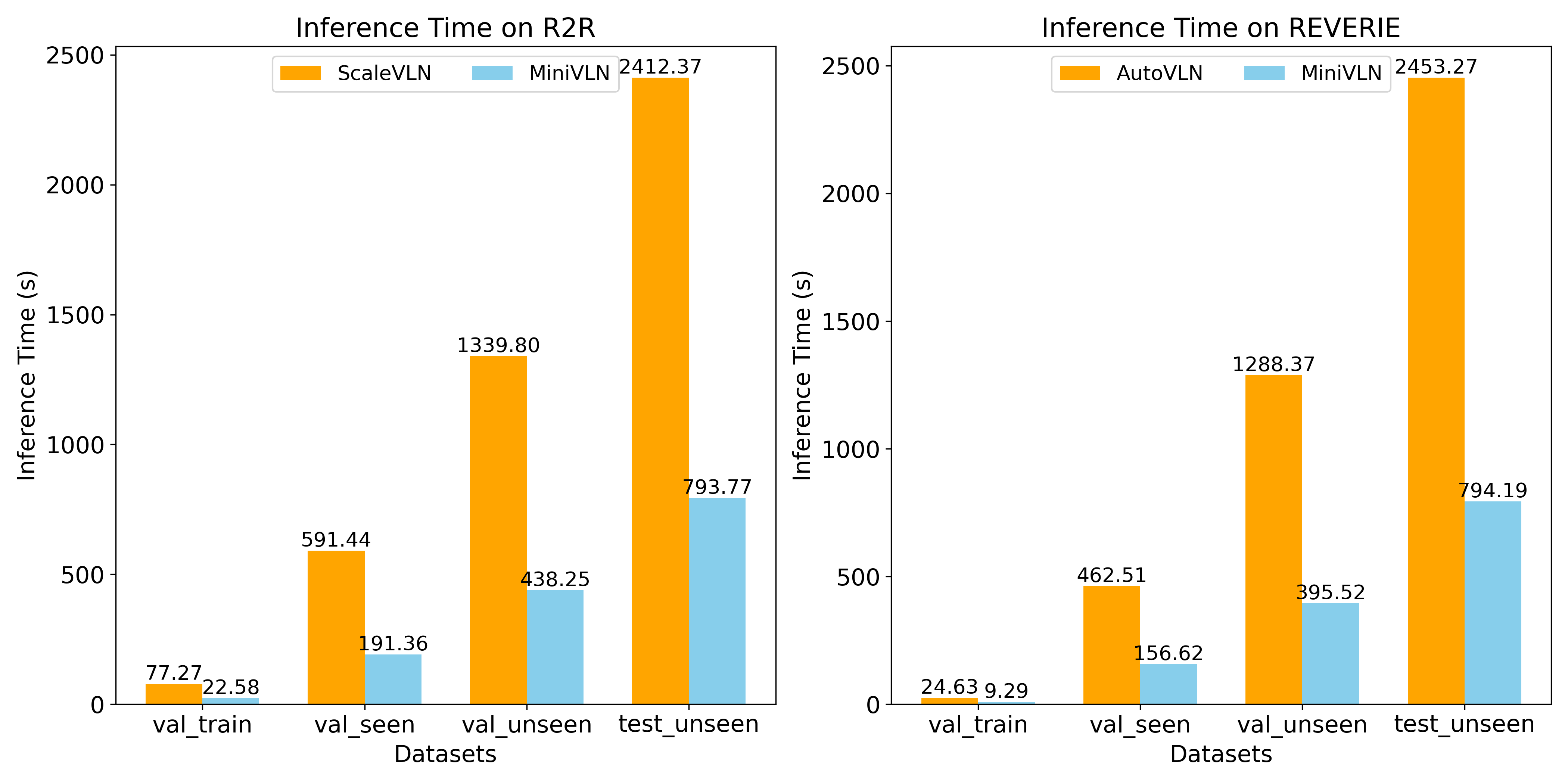}
      \caption{\textbf{The inference time comparison between ScaleVLN and MiniVLN with CPU.} On both datasets, MiniVLN exhibits an inference speed that is more than three times faster than ScaleVLN.}
      \label{fig:speed}
\end{figure}

\subsubsection{Effects of Distillation Objective}
As shown in \Cref{table7}, we conduct an ablation study on the three distillation losses during the fine-tuning stage. The results indicate that distilling knowledge from the fusion of local and global information is most crucial for navigation performance. Additionally, the distillation of textual and visual knowledge also positively impacts overall performance.

\begin{table}[t]
\caption{Ablation on Distillation Objective of the fine-tuninng stage on the R2R dataset.}
\label{table7}
\centering
\begin{tabular}{c|ccc|cc|cc}
\toprule
\multirow{2}{*}{Id}  &  \multicolumn{3}{c}{Ablation} &   \multicolumn{2}{|c|}{Validation Seen} & \multicolumn{2}{c}{Validation Unseen} \\
  & text& pano & fuse & SR & SPL &SR & SPL \\
\midrule
\#1 & \ding{51} & \ding{51} & \ding{51} & \textbf{78.35} & \textbf{70.69}  & \textbf{78.80} & \textbf{70.17}  \\
\hline
\#2 &  & \ding{51} & \ding{51} & 76.79 & 69.61  & 78.46 & 69.62  \\
\#3 & \ding{51} &  & \ding{51} & 76.79 & 69.45  & 77.78 & 69.56  \\
\#4 & \ding{51} & \ding{51} &  & 76 & 69.53  & 75.01 & 67.16  \\
\#5 &  &  & \ding{51} & 77.96 & 70.35  & 78.12 & 69.12 \\
\bottomrule
\end{tabular}
\end{table}

\begin{table}[t]
\caption{Ablation on KDWeight on the R2R dataset.}
\label{table6}
\centering
\begin{tabular}{c|cc|cc}
\toprule
\multirow{2}{*}{KD\-Weight}  & \multicolumn{2}{c|}{Validation Seen} & \multicolumn{2}{c}{Validation Unseen} \\
   &SR & SPL &SR & SPL \\
\midrule
0.01 & 76.30 & 69.49 & 76.76 & 67.89  \\
0.1 & \textbf{78.35} & \textbf{70.69}  & \textbf{78.80} & \textbf{70.17}  \\
1 & 76.98 & 69.78 & 78.20 & 69.94  \\
\bottomrule
\end{tabular}
\end{table}

\subsection{Deployment}
To simulate deployment, we run the complete inference process of the model on the Intel i9-14900HX CPU of a mobile laptop. As shown in \Cref{fig:speed}, the results indicate that MiniVLN achieves over three times the inference speed compared to the teacher model. 

\section{CONCLUSIONS}
In this paper, we aim to enhance the efficiency of VLN models through knowledge distillation, enabling deployment on mobile or edge devices. We propose a progressive two-stage knowledge distillation framework: in the pre-training phase, the model focuses on learning fine-grained knowledge, while in the fine-tuning phase, it learns knowledge that directly impacts navigation decisions. Notably, Our experiments show that the two-stage distillation method enables the student model to more closely match the teacher model's performance than the single-stage approach. Our approach reduces the model size to 12\% of the original, offering a high-performance, low-complexity solution that is particularly well-suited for deployment on resource-constrained devices, such as mobile and edge platforms in embodied VLN scenarios.








\newpage

\bibliographystyle{IEEETranS}
\bibliography{main}

\end{document}